\documentclass{article}

\usepackage{PRIMEarxiv}
\usepackage[utf8]{inputenc} 
\usepackage[T1]{fontenc}    
\usepackage{hyperref}       
\usepackage{url}            
\usepackage{booktabs}       
\usepackage{amsfonts}       
\usepackage{nicefrac}       
\usepackage{microtype}      
\usepackage{lipsum}
\usepackage{fancyhdr}       
\usepackage{graphicx}       
\usepackage{xcolor}
\graphicspath{{media/}}     
\usepackage{multirow}
\title{OCRBench: On the Hidden Mystery of OCR in Large Multimodal Models
}

\author{
  \parbox{\linewidth}{\centering
Yuliang Liu$^{1}$, Zhang Li$^{1}$, Mingxin Huang$^{5}$, Biao Yang$^{1}$, Wenwen Yu$^{1}$, Chunyuan Li$^{2}$,  \\ Xu-Cheng Yin$^{3}$, Cheng-Lin Liu$^{4}$, Lianwen Jin$^{5}$, Xiang Bai$^{1}$\thanks{Corresponding Author.}
  }
  \\
  \parbox{\linewidth}{\centering\vspace{2mm}
  $^{1}$Huazhong University of Science and Technology~~$^2$Microsoft Research, Redmond \\$^3$University of Science and Technology Beijing~~$^4$Chinese Academy of Sciences~~\\$^5$South China University of Technology
  }
 \\
 \parbox{\linewidth}{\centering\vspace{2mm}
 \tt \small ylliu@hust.edu.cn}
 \\
}

\begin{document}
\maketitle

\begin{abstract}
    Large models have recently played a dominant role in natural language processing and multimodal vision-language learning. However, their effectiveness in text-related visual tasks remains relatively unexplored. In this paper, we conducted a comprehensive evaluation of Large Multimodal Models, such as GPT4V and Gemini, in various text-related visual tasks including Text Recognition, Scene Text-Centric Visual Question Answering (VQA), Document-Oriented VQA, Key Information Extraction (KIE), and Handwritten Mathematical Expression Recognition (HMER). To facilitate the assessment of Optical Character Recognition (OCR) capabilities in Large Multimodal Models, we propose \textbf{OCRBench}, a comprehensive evaluation benchmark.
    \textbf{OCRBench} contains 29 datasets, making it the most comprehensive OCR evaluation benchmark available. Furthermore, our study reveals both the strengths and weaknesses of these models, particularly in handling multilingual text, handwritten text, non-semantic text, and mathematical expression recognition. Most importantly, the baseline results presented in this study could provide a foundational framework for the conception and assessment of innovative strategies targeted at enhancing zero-shot multimodal techniques. The evaluation pipeline and benchmark are available at \url{https://github.com/Yuliang-Liu/MultimodalOCR}.
\end{abstract}

\maketitle

\section{Introduction}
The advent of large models has unlocked a wealth of potentials in the realm of advanced computing. In recent years, there has been an explosion in the development of large language models (LLM) such as ChatGPT~\cite{chatgpt} and GPT-4~\cite{gpt4}, giving rise to extraordinary applications in zero-shot task transfer to many new real-world scenarios. 
The success of proprietary LLMs has stimulated tremendous interest in developing open-source LLMs. Among them, LLaMA~\cite{touvron2023llama} is an open-source LLM that matches the performance of GPT-3, followed by Alpaca~\cite{alpaca}, Vicuna~\cite{vicuna}, GPT-4-LLM~\cite{peng2023instruction} to improve the LLM's alignment ability to follow human instruction, reporting impressive performance compared with proprietary LLMs.

The success of large models has also been extended to the multimodal vision-language space~\cite{gan2022vision}, leading to a line of research on large multimodal models (LMM), including contrastive learning~\cite{radford2021learning,yuan2021florence,jia2021scaling,li2022elevater} and generative modeling~\cite{gpt4,driess2023palm,alayrac2022flamingo,wang2022git,liu2023visual}. Surprisingly, Liu et al.~\cite{liu2023visual} show that LMM exhibits excellent zero-shot OCR performance in the wild, without explicitly training on the OCR domain-specific data.
Understanding the efficacy of LMM in handling text-related visual tasks is pivotal, given their potential to infer context from multiple data sources, such as text and images. Despite this advantage, these models may face challenges when dealing with complex relationships between different data types due to their general training on web-scale data. Recognizing these limitations could guide improvements in multimodal methodologies and inspire the creation of more robust models that can handle text-related tasks more efficiently. Additionally, this knowledge can open up novel applications in areas such as digital marketing or social media analysis, where understanding the interplay between textual and visual content in the images is crucial.

To this end, we conduct a comprehensive study on 14 LMMs by evaluating their OCR ability on five representative tasks: Text Recognition, Scene Text-Centric VQA, Document-Oriented VQA, Key Information Extraction, and Handwritten Mathematical Expression Recognition. The results indicate that even state-of-the-art large multimodal models, such as Gemini~\cite{Gemini} and GPT4V~\cite{gpt4v}, still encounter challenges in recognizing blurry text images, handwritten text, multilingual text, and handwritten mathematical expressions. Moreover, we observe that they heavily rely on semantic understanding to recognize words, often favoring common words over random letter sequences. The above findings demonstrate that even the most powerful LMM  still exhibits significant gaps compared to the domain-specific methods in various text-related tasks. Consequently, there exists a promising opportunity to enhance the OCR capabilities of LMMs through domain-specific adaptations and optimizations.

\section{Related Work}
\subsection{Large Multimodal Models} 
The remarkable success of Large Language Models (LLMs) has paved the way for the development of large multimodal models, which combine pretrained visual models with LLMs to enable their visual capabilities. BLIP2~\cite{li2023blip2} introduces the Querying Transformer (Q-Former) as a means to bridge the gap between vision and language models. Flamingo~\cite{alayrac2022flamingo} and OpenFlamingo~\cite{anas_awadalla_2023_7733589} enhance a frozen pretrained LLM by incorporating novel gated cross-attention-dense layers, enabling conditioning on visual inputs. LLaVA~\cite{liu2023visual} pioneers the use of GPT-4 to generate multimodal instruction-following data. Other works, such as \cite{liu2023improved, zhu2023minigpt4}, also focus on aligning the vision module and LLM for improved multimodal understanding. Additionally, \cite{ye2023mplugowl, ye2023mplugowl2} emphasize modality collaboration for image and text. 
LLaVAR~\cite{zhang2023llavar} collects training data with rich text and uses a higher-resolution CLIP as the visual encoder to enhance LLaVA's OCR ability. BLIVA~\cite{hu2023bliva} combines instruction-aware and global visual features to capture richer image information. MiniGPT4V2~\cite{chen2023minigptv2} uses unique identifiers for different tasks when training the model to better distinguish each task instruction effortlessly. UniDoc~\cite{feng2023unidoc} performs unified multimodal instruction tuning on large-scale instruction following datasets and leverages the beneficial interactions among tasks to enhance the performance of each individual task. Docpedia~\cite{feng2023docpedia} directly processes visual input in the frequency domain rather than the pixel space. Monkey~\cite{li2023monkey} enhanced the LMM's ability to perceive details at a low cost by the generated detailed caption and a high-resolution model architecture. TextMonkey\cite{liu2024textmonkeyocrfreelargemultimodal} introduced window attention to strengthen the correlation between different patches and introduced a token resampler to reduce the length of image tokens. 

\subsection{Benchmarks for Large Multimodal Models}
With the ongoing advancements in Large Multimodal Models, the question of how to effectively evaluate their performance has emerged. \cite{duan2024vlmevalkit,lmms_eval2024} have developed effective systems to assess LMMs. \cite{liu2023visual, liu2023mitigating, ye2023mplugowl} introduces GPT-4 or human evaluation to assess the output of LMMs. MMBench~\cite{liu2023mmbench} evaluates LMM using multiple-choice questions across various dimensions of ability. MME~\cite{fu2023mme} measures both perception and cognition abilities on True or False questions. However, their testing on OCR data is limited. Additionally, the True/False question or multiple-choice question cannot accurately assess the OCR's ability to recognize words in an image. In this paper, we conducted an extensive study on various LMMs for five prominent text-related tasks. To facilitate the evaluation of LMMs' OCR ability, we also present OCRBench, a collection of 1000 manually filtered and corrected question-answer pairs on five representative text-related tasks. 

\section{Experiments}
\subsection{Evaluation Metric and Evaluation Dataset}
The responses generated by LMM often include many explanatory terms, so the exact matching approach or Average Normalized Levenshtein Similarity(ANLS)~\cite{Biten2019SceneTV} used in the original dataset are not suitable for evaluating LMM in zero-shot scenarios. We have defined a unified and simple evaluation criterion for all datasets, which is to determine whether the ground truth (GT) is present in the output of the LMM. To reduce false positives, we filter out questions that have answers containing fewer than 4 symbols from all datasets. Additionally, we choose 3000 question instances from some large datasets for testing purposes.

\begin{figure*}[htbp]
\centering
\includegraphics[width=\textwidth]{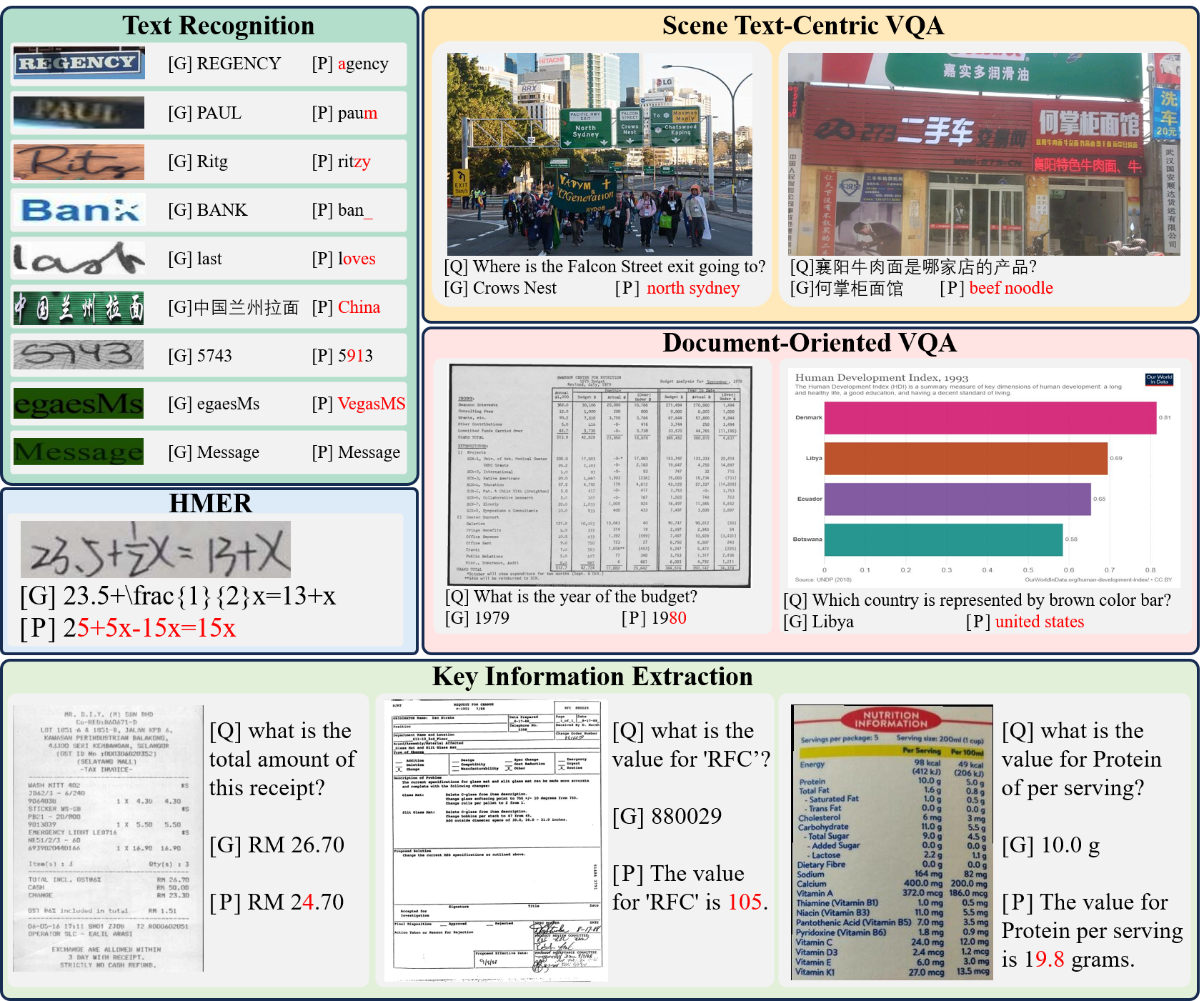}
\caption{Visualization results of the five tasks. `[Q]'  represents question, `[G]' represents ground truth, `[P]' represents prediction generated by LMM.}
\vspace{-1.5em}
\label{fig:all_data}
\end{figure*}

\textbf{Text Recognition}: We evaluate LMM using widely-adopted OCR text recognition datasets, including (1) Regular Text Recognition: IIIT5K~\cite{mishra2012top}, SVT~\cite{Shi2014EndtoendST}, IC13~\cite{Karatzas2013ICDAR2R}; (2)Irregular Text Recognition: IC15~\cite{Karatzas2015ICDAR2C}, SVTP~\cite{Phan2013RecognizingTW}, CT80~\cite{Risnumawan2014ARA}, COCOText(COCO)~\cite{veit2016coco}, SCUT-CTW1500 (CTW)~\cite{Liu2019CurvedST}, Total-Text (TT)~\cite{Chng2017TotalTextAC}; (3) Occlusion Scene Text~\cite{Wang2021FromTT}, encompassing weakly occluded scene text (WOST) and heavily occluded scene text (HOST); (4) Artistic Text Recognition: WordArt~\cite{xie2022toward}; (5) Handwirtten Text Recognition: IAM \cite{marti2002iam}; (6) Chinese Text Recognition: ReCTS~\cite{zhang2019icdar}; (7) Handwritten Digit String Recognition: ORAND-CAR-2014(CAR-A)~\cite{6981115}; (8) Non-Semantic Text(NST) and Semantic Text(ST): LMMs primarily rely on semantic understanding to recognize words. In the experiments, we found that the LMMs have poor recognition performance on character combinations that lacked semantics. To confirm this, we create two datasets: Semantic Text (ST) and Non-Semantic Text (NST) using the IIIT5k dictionary. The ST dataset consists of 3000 images with words from the IIIT5K dictionary, while the NST dataset contains the same words but with shuffled characters without semantics. For English words, we employ the consistent prompt ``What is written in the image''. For Chinese words in ReCTS dataset, we adapt the prompt to ``What are the Chinese characters in the image''. For handwritten digit strings, we utilize the prompt ``What is the number in the image?''.

\textbf{Scene Text-Centric VQA}: We test LMMs on STVQA~\cite{8978179}, TextVQA~\cite{singh2019vqa}, OCRVQA~\cite{8978122} and ESTVQA~\cite{9156857}. Scene Text Visual Question Answering (STVQA) consists of 31,000+ questions across 23,000+ images collected from various public datasets. TextVQA dataset comprises 45,000+ questions on 28,000+ images sampled from specific OpenImages dataset categories expected to contain text. OCRVQA features over 1 million question-answer pairs spanning 207,000+ book cover images. ESTVQA contains 20757 images along with 15056 English questions and 13006 Chinese questions. We have divided the ESTVQA dataset into ESTVQA(CN) and ESTVQA(EN), which specifically include questions and answers in Chinese or English respectively.

\textbf{Document-Oriented VQA}: We assess these LMMs on DocVQA\cite{mathew2021docvqa}, InfographicVQA~\cite{mathew2022infographicvqa} and ChartQA~\cite{masry2022chartqa}. DocVQA is a large-scale dataset with 12,767 document images of diverse types and content, and over 50,000 questions and answers. InfographicVQA is a diverse collection of infographics that includes 5,485 images and a total of 30,035 questions. ChartQA includes a total of 9,608 human-written questions covering 4,804 charts, as well as 23,111 questions generated from human-written chart summaries on 17,141 charts.

\textbf{Key Information Extraction}: We conduct experiments on SROIE~\cite{Huang_2019}, FUNSD~\cite{jaume2019funsd} and POIE~\cite{kuang2023visual}. SROIE contains 1000 complete scanned receipt images for OCR and key information extraction competitions. In this competition, the company, date, address, and total expenditure information must be extracted based on the receipts. FUNSD dataset consists of 199 real, fully annotated, scanned forms that may contain noise. POIE consists of camera images from the Nutrition Facts label of products in English and 3,000 images with 111,155 text instances are collected. The KIE dataset requires the extraction of key-value pairs in the image. To enable LMMs to accurately extract the correct value for a given key in the KIE dataset, we employ manual prompt design. For the SROIE dataset, we utilize the following prompts to assist LMMs in generating the respective values for ``company'', ``date'', ``address'', and ``total'': ``What is the name of the company that issued this receipt?'', ``When was this receipt issued?'', ``Where was this receipt issued?'', and ``What is the total amount of this receipt?''. Additionally, to retrieve the corresponding value for a given key in FUNSD and POIE, we utilize the prompt ``What is the value for `\{key\}'?''
\begin{table}[]
\centering
\resizebox{1\linewidth}{!}{\begin{tabular}{c|ccc|cccccc|cc|c|c|c|c|cc}
\toprule     \multirow{2}{*}{Method}
             & \multicolumn{3}{c|}{Regular}                                        & \multicolumn{6}{c|}{Irregular}                                                                                                           & \multicolumn{2}{c|}{Occluded} & Artistic & Handwritten & Chinese &  Digits & \multicolumn{2}{c}{Semantic}                                                                                                            \\
             & IIIT5K               & SVT                  & IC13            & IC15         & SVTP                 & CT80                 & COCO             & CTW                  & TT             & HOST                 & WOST                  & WordArt              & IAM                  & ReCTS                & ORAND                & NST     & ST         \\ \midrule
BLIP2-6.7B   & 79.1                 & 86.7                 & 83.4                  & 71.2                 & 78.8                 & 76.5                 & 51.4                 & 62.4                 & 68.7                  & 59.8                 & 69.4                  & 68.4                 & 32.9                 & 0                    & 1.2                  & 13.0                   & 82.6                 \\
mPLUG-Owl    & 81.1                 & 84.3                 & 85.9                  & 67.5                 & 73.9                 & 81.4                 & 52.3                 & 69.2                 & 74.4                  & 49.7                 & 62.7                  & 72.1                 & 34.8                 & 0                    & 13.5                 & 44.7                 & 92.3                 \\
InstructBLIP & 86.3                 & 92.0                   & 86.8                  & \color{blue}{80.9}                 & 85.6                 & 86.3                 & 62.6                 & 70.8                 & 77.9                  & \color{blue}{67.8}                 & \color{blue}{78.8}                  & \textbf{73.7}                 & 42.6                 & 0                    & 18.4                 & 25.3                 & 89.7                 \\
LLaVAR       & 84.0                   & 87.6                 & 87.7                  & 79.4                 & 84.0                   & 84.5                 & 61.9                 & 69.5                 & 75.6                  & 61.1                 & 71.9                  & 67.1                 & 49.4                 & 0                    & 9.8                  & 36.2                 & 86.5                 \\
BLIVA        & 86.5                 & \color{blue}{90.6}                 & 87.3                  & \color{blue}{80.9}                 & \color{blue}{87.7}                 & 86.7                 & \textbf{64.8}                 & 71.2                 & 78.1                  & 67.7                 & 77.6                  & \textbf{73.7}                 & 45.1                 & 0                    & 13.8                 & 20.4                 & 89.4                 \\
mPLUG-Owl2   & 80.9                 & 69.6                 & 79.8                  & 53.9                 & 53.5                 & 74.8                 & 52                   & 59.1                 & 60.9                  & 32.5                 & 50.6                  & 60.6                 & 23.8                 & 0                    & 9.9                  & 48.2                 & 93.9                 \\
LLaVA1.5-7B     & 84.2                 & 85.7                 & 86.4                  & 71.9                 & 79.8                 & 82.7                 & 55.6                 & 66.8                 & 73.2                  & 61.4                 & 70.6                  & 68.7                 & \color{blue}{55.4}                 & 0                    & 10.4                 & 15.2                 & 85.3                 \\
UniDoc       & \color{blue}{91.9}                 & 89.2                 & \color{blue}{90.9}                  & 78                   & 80.3                 & \color{blue}{88.2}                 & 64.1                 & \color{blue}{75.3}                 & \color{blue}{78.2}                  & 52.4                 & 68.5                  & -                    & -                    & -                    & -                    & -                    & -                    \\
Monkey       & 83.7                 & 75.1                 & 85.4                  & 53.4                 & 58.4                 & 73.9                 & 43.5                 & 64.5                 & 64.6                  & 43.3                 & 54.9                  & 67.7                 & 30.3                 & \color{blue}{13.1}                 & \color{blue}{29.1}                 & \color{blue}{56.5}                 & \color{blue}{95.5}                 \\ \midrule
Supervised-SOTA         & \multicolumn{1}{c}{\textbf{96.6}} & \multicolumn{1}{c}{\textbf{93.0}} & \multicolumn{1}{c|}{\textbf{96.7}} & \multicolumn{1}{c}{\textbf{85.7}} & \multicolumn{1}{c}{\textbf{89.3}} & \multicolumn{1}{c}{\textbf{89.9}} & \multicolumn{1}{c}{\color{blue}{64.4}} & \multicolumn{1}{c}{\textbf{78.6}} & \multicolumn{1}{c|}{\textbf{80.1}} & \multicolumn{1}{c}{\textbf{73.1}} & \multicolumn{1}{c|}{\textbf{81.6}} & \multicolumn{1}{c|}{\color{blue}{72.5}} & \multicolumn{1}{c|}{\textbf{91.2}} & \multicolumn{1}{c|}{\textbf{94.8}} & \multicolumn{1}{c|}{\textbf{95.5}} & \multicolumn{1}{c}{\textbf{95.4}} & \multicolumn{1}{c}{\textbf{100.0}} \\ \bottomrule
\end{tabular}}
\caption{Text recognition results. Bold black digits indicate the best result, while blue signifies the second best.}
\label{tab:recognition}
\end{table}

\begin{table}[]
\centering
\resizebox{1\linewidth}{!}{\begin{tabular}{c|ccccc|ccc|ccc|c}
\toprule       \multirow{2}{*}{Method}
             & \multicolumn{5}{c|}{Scene Text-Centric VQA}        & \multicolumn{3}{c|}{Document-Oriented VQA.}                    & \multicolumn{3}{c|}{KIE} & HMER    \\
             & STVQA & TextVQA & OCRVQA & ESTVQA(EN) & ESTVQA(CN) & DocVQA & InfoVQA & ChartQA & FUNSD   & SROIE  & POIE  & HME100K \\ \midrule
BLIP2-6.7B   & 20.9  & 23.5    & 9.7    & 40.7       & 0          & 3.2    & 11.3           & 3.4                  & 0.2     & 0.1    & 0.3   & 0       \\
mPLUG-Owl    & 30.5  & 34      & 21.1   & 52.7       & 0          & 7.4    & 20             & 7.9           & 0.5     & 1.7    & 2.5   & 0.1     \\
InstructBLIP & 27.4  & 29.1    & 41.3   & 48.6       & 0.1        & 4.5    & 16.4           & 5.3             & 0.2     & 0.6    & 1     & 0       \\
LLaVAR       & 39.2  & 41.8    & 24     & 58.2       & 0          & 12.3   & 16.5           & 12.2              & 0.5     & 5.2    & 5.9   & 0       \\
BLIVA        & 32.1  & 33.3    & 50.7   & 51.2       & 0.2        & 5.8    & 23.6           & 8.7             & 0.2     & 0.7    & 2.1   & 0.1     \\
mPLUG-Owl2   & 49.8  & 53.9    & 58.7   & \color{blue}{68.6}       & 4.9        & 17.9   & 18.9           & 19.4            & 1.4     & 3.2    & 9.9   & 0       \\
LLaVA1.5-7B     & 38.1  & 38.7    & 58.1   & 52.3       & 0          & 8.5    & 14.7           & 9.3            & 0.2     & 1.7    & 2.5   & 0       \\
UniDoc       & 35.2  & 46.2    & 36.8   & -          & -          & 7.7    & 14.7           & 10.9                    & 1       & 2.9    & 5.1   & \color{blue}{0.4}     \\
DocPedia     & 45.5  & 60.2    & 57.2   & -          & -          & 47.1   & 15.2           & 46.9                 & \color{blue}{29.9}    & 21.4   & \color{blue}{39.9}  & -       \\
Monkey       & \color{blue}{54.7}  & \color{blue}{64.3}    & \color{blue}{64.4}   & \textbf{71}         & \color{blue}{42.6}       & \color{blue}{50.1}   & \color{blue}{25.8}           & \color{blue}{54.0}            & 24.1    & \color{blue}{41.9}   & 19.9 & 0.2     \\ \midrule 
Supervised-SOTA & \textbf{69.6} &\textbf{73.7} & \textbf{68.1} & 43.3 & \textbf{43.3} & \textbf{90.16} & \textbf{36.8} & \textbf{70.5} & \textbf{93.1} & \textbf{98.7} & \textbf{79.5} & \textbf{64.3} \\ \bottomrule
\end{tabular}}
\caption{Results of Scene Text-Centric VQA, Document-Oriented VQA KIE and HMER. Bold black digits indicate the best result, while blue signifies the second best. Since the Supervised-SOTA on ESTVQA dataset does not provide separate results on Chinese and English data, the average performance 42.3 is used as a reference.}
\label{tab:vqa}
\end{table}
\textbf{Handwritten Mathematical Expression Recognition(HMER)}: We evaluate on HME100K~\cite{yuan2022syntax}, which consists of 74,502 images for training and 24,607 images for testing, with 245 symbol classes. During evaluation, we use the prompt ``Please write out the expression of the formula in the image using LaTeX format.''.

\subsection{Results}
The results of text recognition are shown in Tab.~\ref{tab:recognition}. LMMs achieve comparable performance to state-of-the-art supervised models in recognizing regular text, irregular text, occluded text, and artistic text. Particularly in the WordArt dataset, which predominantly comprises challenging artistic text, InstructBLIP2, and BLIVA even outperform the supervised state-of-the-art model. However, LMMs exhibit poor performance in recognizing handwritten text, Chinese text, handwritten strings, and non-semantic text. In tasks such as Scene Text-Centric VQA, Document-Oriented VQA, and KIE, LMMs with smaller input resolutions consistently produce poorer results compared to those with larger input sizes. This is due to the intricate structures and varying sizes of texts, requiring LMMs to capture fine details. The results are shown in Tab.~\ref{tab:vqa}. LMMs face challenges in accurately recognizing handwritten mathematical expressions. By extensively analyzing the results, we provide a qualitative overview of the limitations of LMMs in text-related tasks:
\begin{itemize}
\item \textbf{Semantic-Reliance.} LMMs primarily rely on semantic understanding to recognize words. In our experiments, we observed that LMMs exhibit poor recognition performance when it comes to character combinations that lack semantic meaning. Specifically, when we altered the order of letters in each word, the accuracy of LMMs on the NST dataset decreased by an average of 57.0\% compared to the ST dataset, while the SOTA method for scene text recognition only drops by around 4.6\%. We believe this is because the SOTA method for scene text recognition directly recognizes each character, and semantic information is just used to assist the recognition process, while LMMs primarily rely on semantic understanding to recognize words. This finding is further supported by the low accuracy observed on the ORAND dataset. As shown in Fig.~\ref{fig:all_data}, LMM successfully identified the word ``Message'', but incorrectly recognized ``egaesMs'', which is a rearranged version of the word ``Message''.
\begin{table}[]
\centering
\resizebox{0.8\linewidth}{!}{\begin{tabular}{c|ccccc|c}
\toprule        Method
             & Recog. & $VQA^{S}$ & $VQA^{D}$ & KIE & HMER& Final Score \\ \midrule
Gemini       & \textbf{215}                   & \textbf{174}                         & 128                   & 134      & 8         & \textbf{659}               \\ 
GPT4V        & 167                   & 163                         & \textbf{146}                   & \textbf{160}      & \textbf{9}         & 645               \\
Monkey       & 174                   & 161                         & 91                    & 88       & 0         & 514               \\
mPLUG-Owl2   & 153                   & 153                         & 41                    & 19       & 0         & 366               \\
LLaVAR       & 186                   & 122                         & 25                    & 13       & 0         & 346               \\
LLaVA1.5-13B & 176                   & 129                         & 19                    & 7        & 0         & 331               \\
LLaVA1.5-7B  & 160                   & 117                         & 15                    & 5        & 0         & 297              \\
mPLUG-Owl    & 172                   & 104                         & 18                    & 3        & 0         & 297               \\
BLIVA        & 165                   & 103                         & 22                    & 1        & 0         & 291               \\
InstructBLIP & 168                   & 93                          & 14                    & 1        & 0         & 276               \\
BLIP2   & 154                   & 71                          & 10                    & 0        & 0         & 235               \\
MiniGPT4V2   & 124                   & 29                          & 4                     & 0        & 0         & 157               \\
\bottomrule
\end{tabular}}
\caption{Results of LMMs on OCRBench. Recog. represents text recognition, $VQA^{S}$ represents Scene Text-Centric VQA, $VQA^{D}$ represents Document-Oriented VQA. Bold black digits indicate the best result.}
\label{tab:OCRBench}
\end{table}

\item \textbf{Handwritten Text.} LMMs may encounter difficulties in accurately recognizing handwritten text due to various reasons, including the resemblances in shapes between handwritten letters and numbers. Handwritten text often appears incomplete or blurry due to factors like fast writing speed, irregular handwriting, or low-quality paper. On average, LMMs exhibit a 51.9\% lower performance compared to the supervised state-of-the-art model in this particular task. 

\item \textbf{Multilingual Text.} 
As indicated in Tab.~\ref{tab:recognition} and Tab.~\ref{tab:vqa}, there exists a notable performance gap between ESTVQA(CN) and ESTVQA(EN). The LMMs achieve limited proficiency in the Chinese language. Accurately recognizing Chinese words or responding to Chinese questions poses a considerable challenge for LMMs. Training LMMs with Chinese data emerges as a viable solution. For example, Monkey surpasses other LMMs in Chinese scenarios due to the substantial training of its LLM and visual encoder on Chinese data.

\item \textbf{Fine-grain Perception.} 
Currently, the resolution of most LMMs is limited to 224 x 224 due to the visual encoder used in their architecture. However, supporting high-resolution input is essential for LMMs to capture finer details within images. The restricted input resolution of LMMs like BLIP2 hinders their ability to extract detailed information in tasks such as Scene Text-Centric VQA, Document-Oriented VQA, and KIE. Conversely, LMMs like Monkey, which can handle a resolution of 1344×896, demonstrate enhanced performance in these specific tasks.

\item \textbf{HMER.} 
Recognizing handwritten mathematical expressions poses a challenge for LMMs due to the presence of messy handwritten characters, complex spatial structures, and the indirect LaTeX representation. Additionally, the scarcity of training data for this specific task further complicates the recognition process for LMMs.

\end{itemize}

\begin{figure*}[htbp]
\centering
\includegraphics[width=\textwidth]{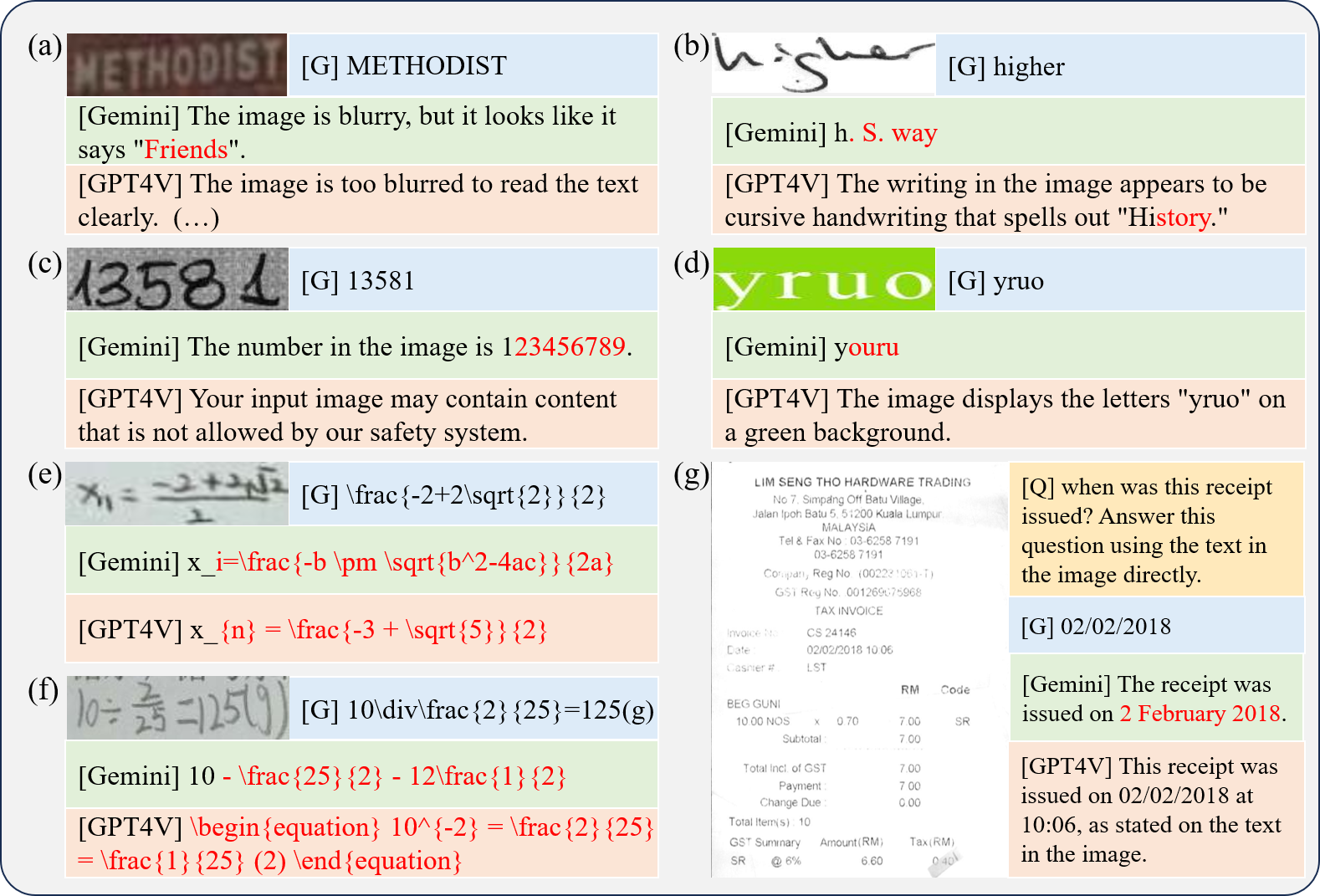}
\caption{Some erroneous results from Gemini and GPT-4V. `[Q]'  represents question, `[G]' represents ground truth. (a) presents the results on blurry image, (b) presents the results on handwritten text, (c) and (d) present the results on Non-Semantic text, (e) and (f) present the results on HMER, (g) display instances where the task instructions are not adhered to.}
\vspace{-1.5em}
\label{fig:GPT4V_Gemini}
\end{figure*}

\subsection{OCRBench}
Evaluating the model on multiple datasets is time-consuming, and the presence of inaccurate annotations in some datasets diminishes the precision of accuracy-based evaluations. To solve these issues, we develop OCRBench to facilitate the accurate and convenient evaluation of LMMs' OCR capabilities. OCRBench consists of five components: text recognition,  Scene Text-Centric VQA, Document-Oriented VQA, KIE, and HMER. It includes 1000 question-answer pairs, and for the KIE task, we added the prompt ``Answer this question using the text in the image directly.'' to restrict the model's response format. The specific composition of OCRBench is shown in the appendix. To ensure a more accurate evaluation, we manually verified and corrected incorrect answers for the 1000 question-answer pairs, providing alternative correct answer candidates. The evaluation results on OCRBench are presented in Tab.~\ref{tab:OCRBench}, where Gemini achieves the highest score, followed by GPT4V in the second position. It is important to note that due to rigorous safety reviews by OPEN AI, GPT4V refused to provide results for 84 images in OCRBench. Monkey exhibited OCR capabilities that trailed only behind GPT4V and Gemini. From Tab.~\ref{tab:OCRBench}, we can observe that even state-of-the-art models like GPT4V and Gemini still struggle with the HMER task. Moreover, they also face challenges in processing unclear images, handwritten text, Non-Semantic text, and adhering to task instructions. As shown in Fig.~\ref{fig:GPT4V_Gemini} (g), even when explicitly requested to answer using the text found in the image, Gemini consistently interprets ``02/02/2018'' as ``2 February 2018''.

\textbf{Why LMMs work for OCR?} While we offer some analysis based on the results, the question as to why these multimodal models can deliver acceptable performance on OCR tasks remains challenging to conclusively explain. 
One plausible explanation for the success of multimodal models in Optical Character Recognition (OCR) tasks lies in the training data of multimodal models (similar to CLIP), which we believe includes some OCR data. However, unsupervised text-image pairs in the training data cannot compete with fully supervised data. 

From the perspective of architecture, the pre-trained visual encoder and Large Language Model (LLM) already demonstrate a solid understanding of their respective domain data, each working within their designated feature spaces. These LLMs connect visual and language data through elements like a linear projection layer, which acts as a visual tokenization step. By aligning visual tokens within the word embedding space of the pre-trained language model, visual embeddings closely resemble their corresponding text embeddings. This alignment facilitates text recognition, allowing the LLM to then present this OCR data to users in a generative manner. Future research could benefit from conducting an ablation analysis to better understand how the volume of multimodal training data impacts OCR performance.
\begin{table}[]
\centering
\resizebox{1.0\linewidth}{!}{\begin{tabular}{llr|llr}
\toprule
Model                       & Affiliation     & \multicolumn{1}{l|}{OCRBench} & Model                        & Affiliation     & \multicolumn{1}{l}{OCRBench} \\ \midrule
MiniCPM-V2.6~\cite{MiniCPM-V2.6}               & OpenBMB & 852                          &
 Cambrian-8B~\cite{tong2024cambrian1fullyopenvisioncentric}                & NYU             & 614                          \\
 InternVL2-Llama3-76B~\cite{chen2024fargpt4vclosinggap}        & Shanghai AI Lab & 842                           &PaliGemma-3B-mix-448~\cite{beyer2024paligemmaversatile3bvlm}        & Google          & 614                          \\
\color{blue}{CongRong}~\cite{CongRong}        & CloudWalk       & 827                           & Cambrian-13B~\cite{tong2024cambrian1fullyopenvisioncentric}                 & NYU             & 610                          \\
 \color{blue}{GLM-4v}~\cite{wang2023cogvlm}              & Zhipu AI        & 814                           & MiniCPM-V-2~\cite{MiniCPM-V-2}                 & OpenBMB         & 605                          \\

MiniMonkey-2B ~\cite{huang2024minimonkeyalleviatesawtootheffect}              & HUST & 802                          &Cambrian-34B~\cite{tong2024cambrian1fullyopenvisioncentric}                 & NYU             & 591                          \\
InternVL2-8B~\cite{chen2024fargpt4vclosinggap}                & Shanghai AI Lab & 794                           & CogVLM-17B-Chat~\cite{wang2023cogvlm}              & Zhipu AI        & 590                          \\
\color{blue}{Claude3.5-Sonnet}~\cite{Claude3.5-Sonnet}    & Anthropic       & 788                           & LLaVA-Next-Yi-34B~\cite{liu2024llavanext}            & UW–Madison      & 574                          \\
\color{blue}{GPT-4o-mini-20240718}~\cite{GPT-4o-mini-20240718} & OpenAI          & 785                           & TextMonkey~\cite{liu2024textmonkeyocrfreelargemultimodal}                   & HUST            & 561                          \\
InternVL2-4B~\cite{chen2024fargpt4vclosinggap}                & Shanghai AI Lab & 784                           & Monkey-Chat~\cite{li2023monkey}                & HUST            & 534                          \\
InternVL2-2B~\cite{chen2024fargpt4vclosinggap}                & Shanghai AI Lab & 781                           & InternLM-XComposer2 ~\cite{dong2024internlmxcomposer2masteringfreeformtextimage}         & Shanghai AI Lab & 532                          \\
GLM-4v-9B~\cite{wang2023cogvlm}                   & Zhipu AI        & 776                           & LLaVA-Next-Vicuna-7B~\cite{liu2024llavanext}         & UW–Madison      & 532                          \\
CogVLM2-19B-Chat~\cite{wang2023cogvlm}             & Zhipu AI        & 757                           & LLaVA-Next-Mistral-7B~\cite{liu2024llavanext}        & UW–Madison      & 531                          \\
InternVL2-1B~\cite{chen2024fargpt4vclosinggap}                 & Shanghai AI Lab & 755                           & \color{blue}{RekaEdge}~\cite{RekaFlash}           & Reka AI         & 506                          \\
\color{blue}{Gemini-1.5-Pro}~\cite{Gemini_models}      & Google          & 754                           & XVERSE-V-13B~\cite{XVERSE-V}                & XVERSE          & 489                          \\
Ovis1.5-Llama3-8B~\cite{lu2024ovisstructuralembeddingalignment}           & Alibaba         & 744                           & Qwen-VL-Chat~\cite{bai2023qwenvlversatilevisionlanguagemodel}                 & Alibaba         & 488                          \\
\color{blue}{Qwen-VL-Plus}~\cite{bai2023qwenvlversatilevisionlanguagemodel} & Alibaba         & 726                           & InternLM-XComposer2-1.8B~\cite{dong2024internlmxcomposer2masteringfreeformtextimage}     & Shanghai AI Lab & 447                          \\
MiniCPM-Llama3-V2.5~\cite{MiniCPM-Llama3-V-2.5}         & OpenBMB         & 725                           & Emu2\_chat~\cite{sun2024generativemultimodalmodelsincontext}                   & BAAI            & 436                          \\
InternVL-Chat-V1.5~\cite{chen2024fargpt4vclosinggap}          & Shanghai AI Lab & 720                           & DeepSeek-VL-7B~\cite{lu2024deepseekvlrealworldvisionlanguageunderstanding}              & DeepSeek        & 435                          \\
\color{blue}{Claude3-Opus}~\cite{Claude3}         & Anthropic       & 694                           & OmniLMM-12B~\cite{OmniLMM-12B}                 & OpenBMB         & 420                          \\
\color{blue}{RekaFlash}~\cite{RekaFlash}            & Reka AI         & 692                           & TransCore-M~\cite{TransCore-M}                 & PCI Research    & 405                          \\
InternLM-XComposer2.5~\cite{zhang2024internlmxcomposer25versatilelargevision}       & Shanghai AI Lab & 686                           & LLaVA-InternLM2-7B~\cite{2023xtuner}          & Shanghai AI Lab & 402                          \\
\color{blue}{Qwen-VL-Max}~\cite{bai2023qwenvlversatilevisionlanguagemodel}          & Alibaba         & 684                           & ShareGPT4V-13B~\cite{chen2023sharegpt4vimprovinglargemultimodal}              & Shanghai AI Lab & 398                          \\
\color{blue}{Gemini-1.0-Pro}~\cite{Gemini_models}       & Google          & 680                           & 360VL-70B~\cite{360VL}                  & QiHoo360        & 397                          \\
InternLM-XComposer2-4KHD~\cite{dong2024internlmxcomposer24khdpioneeringlargevisionlanguage}    & Shanghai AI Lab & 675                           & MiniCPM-V ~\cite{MiniCPM-V}                   & OpenBMB         & 366                          \\
\color{blue}{Claude3-Haiku}~\cite{Claude3}         & Anthropic       & 658                           & Yi-VL-34B~\cite{Yi-VL-34B}                    & 01-AI           & 290                          \\
Mini-InternVL-Chat-2B-V1.5~\cite{chen2024fargpt4vclosinggap}  & Shanghai AI Lab & 652                           & IDEFICS-80B-Instruct~\cite{laurencon2023obelics}         & Hugging Face    & 283                          \\
\color{blue}{Claude3-Sonnet}~\cite{Claude3}       & Anthropic       & 646                           & LLaVA-Next-Llama3~\cite{liu2024llavanext}            & ByteDance       & 252                          \\
Mini-InternVL-Chat-4B-V1.5~\cite{chen2024fargpt4vclosinggap}  & Shanghai AI Lab & 639                           & MMAlaya~\cite{datacanvas2024mmalaya}                      & DataCanvas      & 223                          \\
Phi-3-Vision~\cite{Phi-3-Vision}                & Microsoft       & 637                           & VisualGLM~\cite{du2022glm}                   & Tsinghua        & 170                          \\
WeMM~\cite{WeMM}                    & WechatCV        & 628                           & OpenFlamingo v2~\cite{awadalla2023openflamingo}              & UW              & 149                          \\
IDEFICS2-8B~\cite{laurençon2024matters}                 & Hugging Face    & 626                           & PandaGPT-13B~\cite{su2023pandagptmodelinstructionfollow}                & Tencent AI Lab  & 36                           \\
\color{blue}{Step-1V}~\cite{Step-1V}          & StepFun         & 625                           & Chameleon-30B~\cite{chameleonteam2024chameleonmixedmodalearlyfusionfoundation}                & Meta            & 27                           \\ \bottomrule
\end{tabular}}
\caption{The results of OCRBench across various models are sourced from ~\cite{duan2024vlmevalkit} and their papers. Closed-source models are indicated in blue.}
\label{tab:leaderboard}
\end{table}

\textbf{Future Directions:} Although current LMMs have shown promising results, they still face challenges when dealing with complex tasks and are unable to match domain-specific methods in traditional text-related tasks. To enhance their performance, these models should concentrate on improving their ability to detect intricate features in images, enhancing their recognition of individual character shapes, and developing multilingual datasets for training.

Recent work such as the ICL-D3IE~\cite{He2023ICLD3IEIL}, introduces an in-context learning framework leveraging large language models like GPT-3 and ChatGPT for document information extraction. This approach effectively navigates the modality and task gaps in this field, extracting challenging and distinct segments from difficult training documents, designing relationship demonstrations for enhanced positional understanding, and incorporating formatting demonstrations for answer extraction. Despite its notable success across various datasets and settings, ICL-D3IE encounters challenges in the in-domain setting on CORD~\cite{Park2019CORDAC}. This result points to the potential of large language models in managing tasks involving visually intricate documents and encourages the development of novel, minimally supervised OCR techniques.

Moreover, a study by Li \textit{et al.}~\cite{Li2023EvaluatingCI}, evaluates ChatGPT across seven detailed information extraction tasks. This evaluation reveals that ChatGPT performs well in OpenIE settings, generates high-quality responses, shows a tendency towards overconfidence, and maintains strong fidelity to original texts. However, this study focused exclusively on pure text, without venturing into visually rich document texts within the OCR domain. Future research should therefore explore ChatGPT's performance in these visually rich contexts.

The application of LMMs in specialized domains is also a crucial field, and recent developments have shown their vast potential. One such model, developed by Google, is Med-PaLM2 \cite{singhal2023expertlevel}, which is fine-tuned on medical knowledge from PaLM2. It is the first language model to perform an expert level on US Medical Licensing Examination (USMLE) style questions and can analyze patients' conditions through medical images, including plain films and mammograms. Google claims that it has approached the performance of clinician experts. Additionally, LLaVA-Med \cite{li2023llavamed} attempts to extend multimodal instruction-tuning to the biomedical domain and demonstrates excellent chat abilities with domain knowledge. These studies highlight the potential of LMMs in vertical domain tasks. Further exploration of their applications in other domains, such as gaming and education, is also warranted.

Ultimately, these developments and future research directions could potentially pave the way for multimodal models that can more efficiently handle complex tasks like OCR, expanding the application range of LMM.

\section{Future work}
With the advancement of large multimodal models, numerous research institutions have made significant strides in optical character recognition (OCR) capabilities. For instance, MiniCPMV-2.6 achieved a score of 852 on OCRBench, while the Mini-Monkey scored 802 using just 2B parameters. These results underscore the immense potential of large multimodal models in the OCR domain, as highlighted in Table~\ref{tab:leaderboard}. OCRBench has been instrumental in advancing this field. However, it is important to note that the data utilized by most models is not entirely open-source, and many models remain proprietary. This demonstrates an existing opportunity for the open-source community to further explore and enhance the OCR capabilities of large multimodal models.

Furthermore, OCRBench currently lacks a comprehensive coverage of image data types and tasks. For instance, it falls short in encompassing more challenging data types like multilingual documents and texts captured in diverse scenarios, as well as tasks related to text detection. To address these gaps, our future work will focus on expanding OCRBench to include these elements, thereby fostering the ongoing advancement and utilization of large multimodal models in the realm of OCR.

\section{Conclusion}
This paper has presented an extensive study on the performance of LMM on OCR tasks, including text recognition, Scene Text-Centric, Document-Oriented VQA, KIE, and HMER. Our quantitative assessment reveals that LMM can achieve promising results, especially in text recognition, even attaining SOTA performance in some datasets. However, significant gaps persist compared to domain-specific supervised methods, suggesting that specialized techniques tailored to each task are still essential, as the latter uses much less computational resources and data.
The proposed OCRBench has served as the evaluation benchmark for the OCR capabilities of multimodal large models, driving the development of LMMs and demonstrating their immense potential in the OCR field. In the future, we hope to further explore the potential of LMMs across more scenarios, more complex tasks, and multiple languages, ultimately paving the way for more intelligent and versatile OCR solutions for large multimodal models.

\section{Acknowledgements}
This work was supported by National Natural Science Foundation of China (Grant Nos. 62225603 and 62226104).Thanks for the help from Hongliang Li, Yang Liu, Dezhi Peng, Mingyu Liu and Mingrui Chen.

\bibliographystyle{unsrt}  
\bibliography{references}  
\newpage
\appendix
\section{Summary of the Evaluation Benchmarks.}

Tab.~\ref{tab:detailOCR} presents the detailed composition of OCRBench. The Text Recognition task consists of a total of 300 images, including Regular Text Recognition, Irregular Text Recognition, Artistic Text Recognition, Handwriting Recognition, Digit String Recognition, and Non-Semantic Text Recognition, with 50 images each. The Scene Text-centric VQA task includes 200 questions, with 50 questions each from STVQA, TextVQA, ESTVQA(EN), and OCRVQA datasets. The Doc-oriented VQA task comprises 200 questions, with 50 questions each from DocVQA, ChartQA(Aug.), ChartQA(Hum.), and InfoVQA datasets. The Key Information Extraction task consists of 200 questions, with 67 questions from SROIE, 66 questions from FUNSD, and 67 questions from POIE. The Handwritten mathematical expression recognition task includes 100 images from the HME100k dataset. OCRBench encompasses a total of 1000 questions across these five tasks, and all answers have been manually filtered and corrected.

\begin{table}[h]
\centering
\begin{tabular}{c|c|c}
\hline
Task                                        & Dataset                       & Smaples \\ \hline
\multirow{6}{*}{Text Recognition}           & Regular Text Recognition      & 50             \\
                                            & Irregular Text Recognition    & 50             \\
                                            & Artistic Text Recognition     & 50             \\
                                            & Handwriting Recognition       & 50             \\
                                            & Digit String Recognition      & 50             \\
                                            & Non-Semantic Text Recognition & 50             \\ \hline
\multirow{4}{*}{Scene Text-centric VQA}     & STVQA                         & 50             \\
                                            & TextVQA                       & 50             \\
                                            & ESTVQA(EN)                        & 50             \\
                                            & OCRVQA                        & 50             \\ \hline
\multirow{4}{*}{Doc-oriented VQA}           & DocVQA                        & 50             \\
                                            & ChartQA(Aug.)                 & 50             \\
                                            & ChartQA(Hum.)                 & 50             \\
                                            & InfoVQA                       & 50             \\ \hline
\multirow{3}{*}{Key Information Extraction} & SROIE                         & 67             \\
                                            & FUNSD                         & 66             \\
                                            & POIE                          & 67             \\ \hline
HMER                                        & HME100k                       & 100            \\ \hline
All                                         & -                             & 1000            \\ \hline
\end{tabular}
\caption{Details of the OCRBench.}
\label{tab:detailOCR}
\end{table}

\section{Summary of the Models.}
Tab.~\ref{tab:detail_models} displays the details of the testing models, with most LMMs having their resolution limited to an input size of 224 or 336 due to the visual module. As a result, they perform poorly on Doc-oriented VQA and KIE tasks. Conversely, models with higher resolutions exhibit better performance on these tasks. 

\newpage

\begin{table}[]
\centering
\begin{tabular}{c|cc}
\hline
Models       & \multicolumn{1}{c|}{Language Model} & Input Resolution \\ \hline
BLIP2-OPT-6.7B   & \multicolumn{1}{c|}{OPT-6.7B~\cite{zhang2022opt}}       & 224              \\
mPLUG-Owl    & \multicolumn{1}{c|}{LLaMA-2 7B~\cite{touvron2023llama2}}     & 224              \\
InstructBLIP & \multicolumn{1}{c|}{Vicuna-7B~\cite{vicuna}}      & 224              \\
LLaVAR       & \multicolumn{1}{c|}{Vicuna-7B~\cite{vicuna}}      & 336              \\
BLIVA        & \multicolumn{1}{c|}{Vicuna-7B~\cite{vicuna}}      & 224              \\
mPLUG-Owl2   & \multicolumn{1}{c|}{LLaMA-2 7B~\cite{touvron2023llama2}}     & 448              \\
LLaVA1.5-7B  & \multicolumn{1}{c|}{Vicuna-7B~\cite{vicuna}}      & 336              \\
LLaVA1.5-13B & \multicolumn{1}{c|}{Vicuna-13B~\cite{vicuna}}     & 336              \\
MiniGPT4V2   & \multicolumn{1}{c|}{Llama2-Chat-7B~\cite{touvron2023llama2}} & 224              \\
Monkey       & \multicolumn{1}{c|}{Qwen-7B~\cite{bai2023qwen}}        & 896              \\
Unidoc       & \multicolumn{1}{c|}{Vicuna~\cite{vicuna}}         & 336              \\
DocPedia     & \multicolumn{1}{c|}{Vicuna~\cite{vicuna}}         & 2560             \\ \hline
GPT4V        & \multicolumn{2}{c}{gpt-4-vision-preview}               \\ \hline
Gemini       & \multicolumn{2}{c}{gemini-pro-vision}                  \\ \hline
\end{tabular}
\caption{Details of the models.}
\label{tab:detail_models}
\end{table}

\section{Supervised SOTA}
State-of-the-art (SOTA) in widely-adopted OCR text recognition datasets is achieved by PARSeq \cite{bautista2022scene}, which utilizes Permutation Language modeling for enhanced contextual information and unifies context-free non-AR and context-aware AR inference. We train Parseq on the ST \cite{gupta2016synthetic} and MJ \cite{jaderberg2014synthetic} synthetic datasets, and test it directly on real datasets. 
The ReCTS dataset is not commonly used, therefore, we chose TPS-ResNet, which has the highest ranking on the ICDAR ranking table along with a published article, as the SOTA. The model employs Thin-plate-spline (TPS) based Spatial Transformer Network (STN) to normalize the input text images, followed by a ResNet-based feature extractor and BiLSTM for text recognition. Although the evaluation metric for ReCTS is Normalized Edit Distance, we still use it as a reference. AttentionHTR \cite{kass2022attentionhtr} proposes an attention-based sequence-to-sequence model to achieve SOTA on IAM. The word error rate of this method on IAM is 8.76, so we use 91.24 as the corresponding word accuracy. Yu \cite{yu2022efficient} achieves SOTA on ORAND-CAR-2014 by proposing an efficient text line recognition method based on prototype learning
with feature-level sliding windows for classification.

For Scene Text Visual Question Answering (STVQA) and Optical Character Recognition Visual Question Answering (OCR-VQA), SOTA is reached by GIT \cite{wang2022git}. GIT trains a Generative Image-to-text Transformer which contains one image encoder and one text decoder under a single language modeling task.
For TextVQA, Mia \cite{qiao2021winner} achieves SOTA by using T5 \cite{raffel2020exploring} for TextVQA task. To align object feature and scene text, Mia is pretrained by masked language modeling(MLM) and relative position prediction(RPP) task. For DocVQA Dataset, BAIDU-DI assembles ERNIE-Layout~\cite{Peng2022ERNIELayoutLK} and DocPrompt (a few-shot model using multi-stage training based on ERNIE-Layout) to achieve SOTA. ESTVQA dataset has not been explored well, Fang \cite{fang2022cross} achieves SOTA by using a cross-modal attention network to guide the expression of visual and textual features simultaneously while using a transformer decoder to output the results. However, since Fang \cite{fang2022cross} does not provide separate results on Chinese and English data, the average performance is used as a reference. For InfographicVQA, SOTA is reached by DUBLIN\cite{aggarwal2023dublin}. For ChartQA, Liu\cite{liu2022deplot} achieves SOTA by designing a modality conversion module DEPLOT.

The SOTA on the FUNSD dataset is achieved by ERNIE-Layout \cite{Peng2022ERNIELayoutLK}, which enhances layout knowledge by correcting the reading order in pretraining phase. For SROIE, StrucTexT \cite{Li2021StrucTexTST} achieves the SOTA. StrucTexT introduces a segment-token aligned encoder for transformer and is pretrained by Masked Visual Language Modeling task and the new Sentence Length Prediction and Paired Boxes Direction tasks. For POIE, Kuang \cite{kuang2023visual} achieves SOTA by adopting contrastive learning to effectively establish the connections between the tasks of OCR and information extraction. For HME100K, Li \cite{li2022counting} proposes a Counting-Aware Network which jointly optimizes HMER and symbol counting tasks to achieve SOTA.
\end{document}